\documentclass[letterpaper, 10 pt, conference]{ieeeconf}  %

\IEEEoverridecommandlockouts    %
\usepackage{graphics} %
\usepackage{graphicx}
\usepackage{epsfig} %
\usepackage{times} %
\usepackage{amsmath} %
\usepackage{amssymb}  %
\usepackage{float}

\usepackage{booktabs}
\usepackage{multirow}

\usepackage{xcolor}
\usepackage{colortbl}

\usepackage{pifont}%

\newcommand{\cmark}{\ding{51}}%
\newcommand{\xmark}{\ding{55}}%

\usepackage{siunitx}  
\usepackage{subcaption}
\usepackage{lipsum} %

\usepackage[colorlinks = true,
            linkcolor = blue,
            urlcolor  = blue,
            citecolor = blue,
            anchorcolor = blue]{hyperref}

\usepackage{tikz}
\usepackage{leftidx}

\usepackage{bm}

\usepackage[normalem]{ulem} %
\usetikzlibrary{shapes.geometric, arrows}

\tikzstyle{startstop} = [rectangle, rounded corners, minimum width=3cm, minimum height=0.1cm,text centered, draw=black]
\tikzstyle{arrow} = [thick,->,>=stealth]

\usepackage{bm}
\usepackage{siunitx} 

\usepackage{mathtools}

\usepackage{titlesec}
\titlespacing*{\section}{0pt}{1.2ex plus 0.2ex minus 0.1ex}{0.8ex plus 0.1ex}
\titlespacing*{\subsection}{0pt}{1.0ex plus 0.2ex minus 0.1ex}{0.6ex plus 0.1ex}
\titlespacing*{\subsubsection}{0pt}{0.8ex plus 0.2ex minus 0.1ex}{0.5ex plus 0.1ex}

\newcommand{\added}[1]{\textcolor{black}{#1}}
\newenvironment{addedstuff}{\par\color{black}}{\par}
\newcommand{\addedtable}[1]{\color{black}}

\usepackage[utf8]{inputenc}

\usepackage{tikz}
\usepackage{tikz-cd}
\usepackage{pgfplots}
\pgfplotsset{compat=1.14}
\usepackage{ulem}
\usepackage{acro}
\usepackage{tabularx} %

\usepackage{hyperref}

\usepackage{etoolbox}
\makeatletter
\patchcmd{\@maketitle}{\vskip0.25in}{}{}{\PackageWarning{main}{Could not remove the space above the title}}
\makeatother

\newif\ifanonymous
\anonymousfalse
\newcommand{\confpaper}{\ifanonymous the conference paper~\cite{andreu2025focianon}\else our conference paper~\cite{andreu2025foci}\fi}

\title{\LARGE \bf FOCI: Trajectory Optimization on
Gaussian Splats \added{for Gaussian Robots}}

\ifanonymous
\author{\added{Anonymous Author(s)}}
\else
\author{Mario Gomez Andreu*$^{1}$, \added{Carlos Martinez Cuenca*$^{2}$},  Maximum Wilder-Smith*$^{1}$, Victor Klemm$^{1}$, \\ Vaishakh Patil$^{1}$, Marco Hutter$^{1}$, Jesus Tordesillas$^{2}$
\thanks{*These authors contributed equally}
\thanks{$^{1}$The authors are with the Robotic Systems Lab, ETH Z\"urich, Switzerland.\tt{ \{margomez, mwilder, vklemm, patilv, mahutter\}@ethz.ch}}
\thanks{$^{2}$The authors are with the Trustworthy Robotics and AI Lab (TRAIL), IIT, Comillas (Spain). \tt{\added{carlosmartinezcuenca@alu.comillas.edu}, jtordesillas@comillas.edu}}
\thanks{This work was funded by NCCR Automation, and Swiss Federal Railways (SBB) via ETH Mobility Initiative.}}
\fi

\DeclareAcronym{3dgs}{
short = 3DGS ,
long = 3D Gaussian Splatting,
sort = 3DGaussianSplat
}

\DeclareAcronym{nerf}{
short = NeRF ,
long = Neural Radiance Field,
short-plural = s ,
long-plural = s ,
sort = NeuralRadianceField
}

\DeclareAcronym{esdf}{
short = ESDF ,
long = Euclidean Signed Distance Field,
short-plural = s ,
long-plural = s ,
sort = EuclideanSignedDistanceField
}

\makeatletter
\providecommand{\bstctlcite}[1]{\@bsphack
  \@for\@citeb:=#1\do{%
    \edef\@citeb{\expandafter\@firstofone\@citeb}%
    \if@filesw\immediate\write\@auxout{\string\citation{\@citeb}}\fi}%
  \@esphack}
\makeatother

\begin{document}
\bstctlcite{BSTcontrol}
\maketitle
\thispagestyle{empty}
\pagestyle{empty}

\begin{abstract}
\ac{3dgs} has recently gained popularity as a faster alternative to \acp{nerf} in 3D reconstruction and view synthesis methods. 
Leveraging the spatial information encoded in \ac{3dgs}, 
this work proposes FOCI (Field Overlap Collision Integral), an algorithm that is able to optimize trajectories directly on the Gaussians themselves. FOCI leverages a novel and interpretable collision formulation for \ac{3dgs} using the notion of the overlap integral between Gaussians. 
\added{The formulation presented applies to different robot morphologies (e.g., fixed and mobile base manipulators, legged robots, UAVs) and handles both static and dynamic Gaussian-Splat environments.}
Contrary to other approaches, which represent the robot with conservative bounding boxes that underestimate the traversability of the environment, we propose to represent the environment \textit{and} the robot as Gaussian Splats. This not only has desirable computational properties, but also allows for orientation-aware planning, allowing the robot to pass through very tight and narrow spaces.    
We extensively test our algorithm in both synthetic and real Gaussian Splats, showcasing that collision-free trajectories for \added{legged, aerial and manipulator robots} can be computed in seconds, even with hundreds of thousands of Gaussians making up the environment. 
\end{abstract}

\begin{addedstuff}
\urlstyle{same}%
\section*{Supplementary Material}
\ifanonymous
\noindent\textbf{Video}: \url{https://doi.org/10.6084/m9.figshare.35108173}\\
\textbf{Code}: \url{https://anonymous.4open.science/r/foci-D835/}
\else
\noindent\textbf{Video}: \url{https://youtu.be/H8clpnsGk8s}\\
\textbf{Code}: \url{https://github.com/leggedrobotics/foci}
\fi
\end{addedstuff}

\section{INTRODUCTION}
Trajectory planning is integral to autonomous mobile robotics to ensure guided and safe operation.
However, in order to make an informed decision, these planning algorithms heavily depend on the underlying environment representations.
Popular representations include occupancy grids, signed distance fields, 3D Meshes, and point clouds. 

Recently, \acp{nerf}~\cite{Mildenhall2020} have been proposed as a novel neural representation of the environment. However, they suffer from having slow inference speeds because new views have to be created using a computationally expensive ray-casting procedure. More recently, \ac{3dgs}~\cite{Kerbl2023} has been proposed as a promising alternative to \acp{nerf}. Instead of using a neural network to represent the radiance field, it is defined explicitly by a set of 3D Gaussian ellipsoids. Because of this, new views can be efficiently generated by projecting these ellipsoids into the view plane.  

\begin{figure}[t]
    \centering
    \includegraphics[width=\columnwidth]{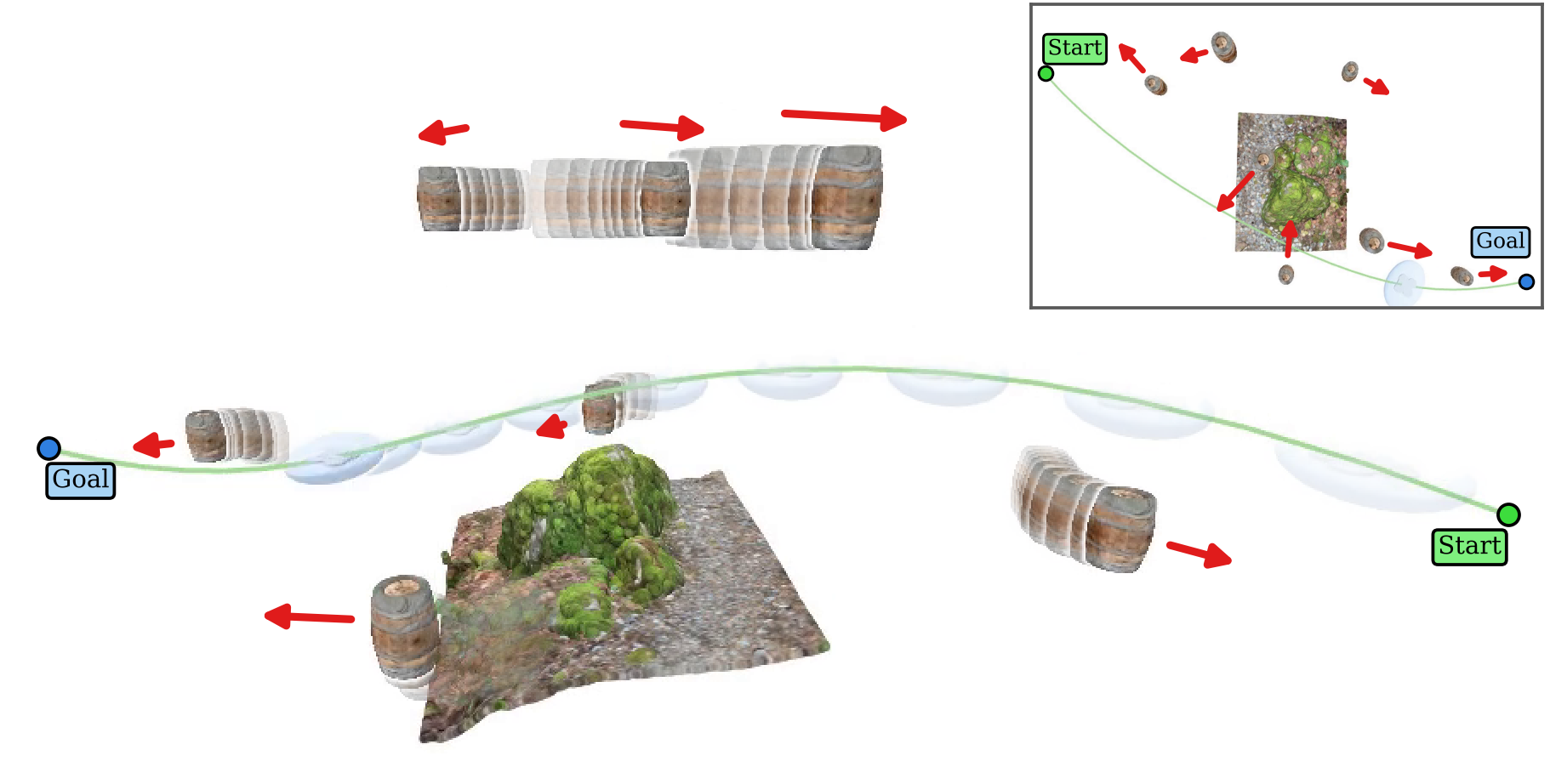}
    \caption{\added{An underactuated UAV (blue Gaussians) plans a collision-free trajectory (green) from Start to Goal, avoiding a static rock ($\sim$1.35M Gaussians) and seven moving barrels, each modeled with a single Gaussian (red arrows indicate their motion direction). Inset: top-down view.}}
    \label{fig:mobile_example}
\end{figure}

\begin{figure}[t]
	\centering
	\includegraphics[width=0.8\columnwidth]{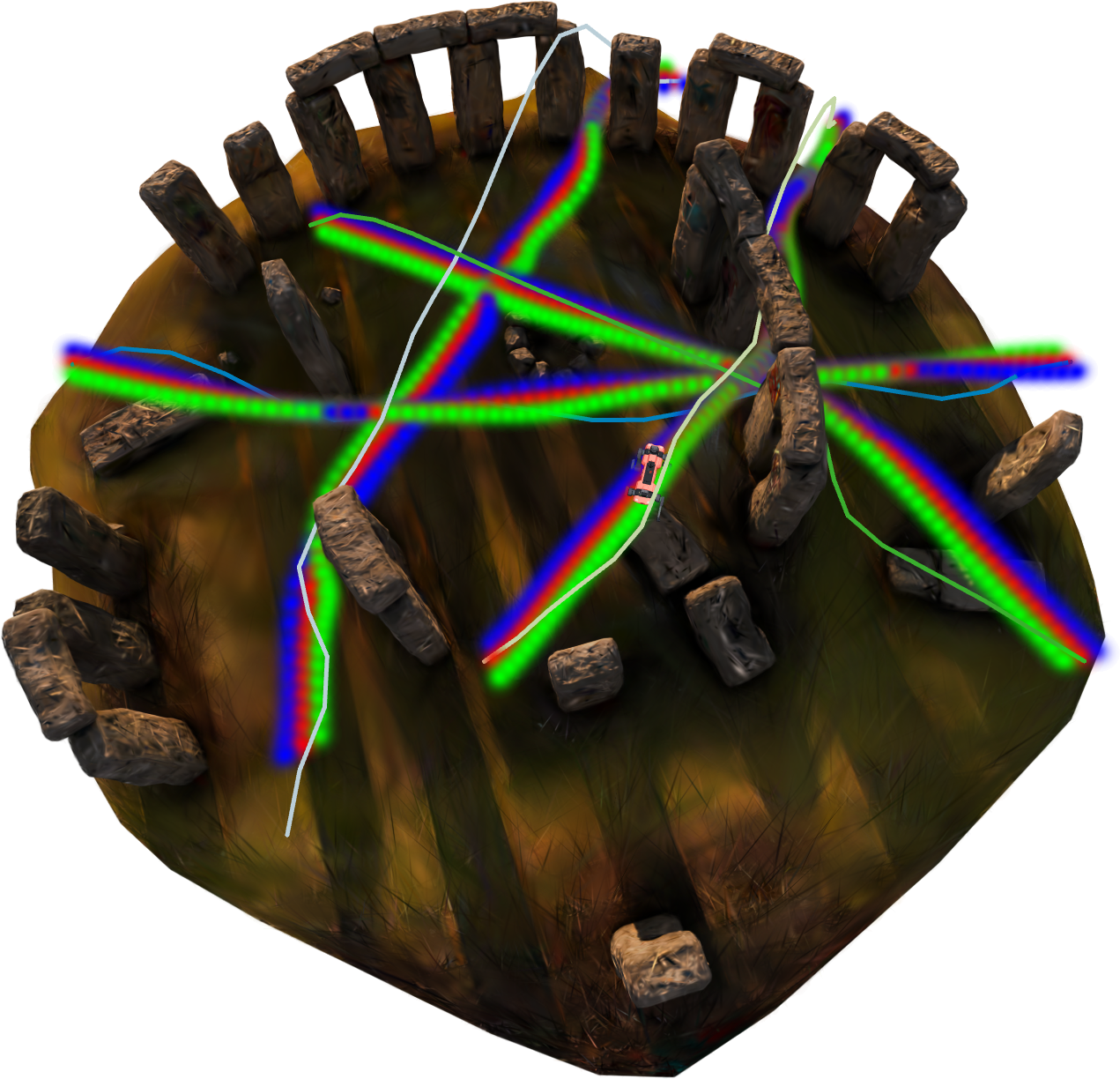}
	\caption{Four orientation-aware trajectories planned through a Stonehenge environment for an ANYmal quadruped robot~\cite{hutter2016anymal}. Splines show A* initial trajectories\added{; robot Gaussians are shown in \textcolor{green}{green} (front), \textcolor{red}{red} (center), and \textcolor{blue}{blue} (rear).}}
	\label{fig:ANYmal_title}
	\vspace*{-0.0cm}
\end{figure}

Given the advantages that \ac{3dgs} offers when compared with \acp{nerf}, the natural question is how a robot can leverage this Gaussian representation for navigation. \textbf{In this paper, we propose an algorithm that enables a robot to perform trajectory optimization directly on the 3D Gaussians.} \added{Furthermore, the proposed formulation is robot-agnostic and makes no assumption of a static environment, enabling trajectory optimization across a wide range of robotic platforms and in dynamic scenarios (see Figs.~\ref{fig:mobile_example} and~\ref{fig:ANYmal_title}).} Although some steps have been taken in this direction~\cite{Chen2024,lei2025gaussnav,goel2025distance, marticorena2025remosplat}, the huge number of Gaussians a scene can have, together with the specific formulation of an explicit collision measure, makes this problem especially hard.

To overcome these challenges, we propose \textbf{FOCI}, a trajectory optimization algorithm that leverages the overlap integral -- the spatial integral over the multiplication of two functions -- as a proxy measure for the collision between two Gaussians. By representing both the robot \textit{and} the environment with 3D Gaussians, evaluating the full-body collision between them reduces to a sum of normal distribution evaluations. Furthermore, the resulting expression is fully differentiable, yielding expressive gradients in the optimization step.

\added{This work extends \confpaper{}, and its contributions are summarized as follows:}
\begin{itemize}
    \setlength{\itemsep}{1pt}
    \setlength{\parskip}{0pt}
    \setlength{\topsep}{2pt}
    \item A novel collision measure between Gaussian Splats based on the overlap integral between Gaussians.
    \item \added{A robot-agnostic, fully differentiable trajectory optimization pipeline compatible with arbitrary robot morphologies and dynamic environments.}
    \item A GPU implementation of the trajectory optimization. 
    \item An evaluation of the proposed algorithm on the ANYmal legged robot in simulation and the real world\added{, and on manipulator and aerial robots in simulation}.
\end{itemize}
\added{In particular, the new contributions of this version with respect to \confpaper{} are:}
{\renewcommand{\labelenumi}{\alph{enumi})}
\begin{enumerate}
    \setlength{\itemsep}{1pt}
    \setlength{\parskip}{0pt}
    \item \added{\textit{Formulation:} extension to arbitrary robot morphologies (fixed and mobile-base manipulators, legged robots, and UAVs) by optimizing the robot joint trajectories, and to dynamic environments through time-varying environment Gaussians.}
    \item \added{\textit{Experiments:} new experiments with a manipulator and UAVs, including a UAV navigating among moving obstacles.}
    \item \added{\textit{Comparisons:} benchmarks against the trajectory optimizers CHOMP~\cite{zucker2013chomp} and STOMP~\cite{kalakrishnan2011stomp}.}
\end{enumerate}}

\section{Background and Related Work}

\subsection{Radiance Fields}
Radiance Fields are widely used in visual computing to represent 3D scenes and perform novel view synthesis.
The preliminary form is a \acl{nerf}~\cite{Mildenhall2020}, which provides a neural mapping from position in space and a viewing angle to visual density and color along a ray.
As an alternative, \ac{3dgs}~\cite{Kerbl2023} produces similar quality reconstruction with an explicit basis at far faster speeds. \ac{3dgs} creates a series of 3D Gaussian ellipsoids with colors and Spherical Harmonics~\cite{fridovich2022plenoxels} to encode view-dependent effects. Instead of ray casting each Gaussian, the ellipsoids are splatted~\cite{964490} onto the viewing plane for rasterization levels of speed.

\subsection{Trajectory Optimization}

\subsubsection{Standard Methods}

Trajectory planning generates a path for a robot that is collision-free with the surrounding environment and further minimizes a performance metric such as execution time or control efforts.
Depending on the formulation, trajectory optimization also includes optimizing control inputs to follow the trajectory. Similar to other works~\cite{Chen2024,lei2025gaussnav,jin2024gs}, we assume robust low-level control that can track reasonably smooth state trajectories, so we omit the control inputs in our optimization formulation.

Graph-based methods such as A*~\cite{astar}, and sampling-based methods such as RRT*~\cite{karaman2011anytime} are straight-forward to implement and produce a series of collision-free waypoints. Trajectories produced by graph-based methods are often good initial guesses for other approaches. 
Unlike graph- or sampling-based methods, spline-based methods optimize a continuous trajectory instead of a series of waypoints. Collision avoidance is ensured by either constraining the convex hull of this spline to a predetermined safety corridor or by ensuring no collision at discretized points along the spline.
Zhou~et~al.~\cite{Zhou2020} obtain an obstacle gradient to push spline trajectories out of obstacles to achieve collision-free trajectories while Gao et al.~\cite{gao2016,Gao2019} construct a safe flight corridor around an RRT* path through a point cloud and optimize a spline trajectory inside the corridor to ensure safety.

\subsubsection{Radiance Field Methods}
There are numerous trajectory optimizing algorithms designed to work on standard scene representations such as meshes or point clouds, and now -- with the recent emergence of Radiance Fields -- \acp{nerf} are being used to tackle the problem.
Planning directly on Radiance Fields can be favorable as they are often trained from monocular images, and the extensive reconstruction allows for their use in multiple downstream tasks from planning and SLAM to teleoperation.
Pantic~et~al.~\cite{Pantic2022} propose a method to obtain an \ac{esdf} directly from a \ac{nerf} by inserting an additional head into the neural network to output the encoded distance. Adamkiewicz et al.~\cite{adamkiewicz2022vision} directly use the opacity output by the \ac{nerf} as a collision measure at a set of points of the robot body. Starting from an A*-initial guess, a joint cost function consisting of a control effort and opacity term at the control points is minimized. 
In CATNIPS~\cite{Chen2023}, a probabilistic approach is used and shows that a \ac{nerf} is equivalent to a Poisson Point Process, which allows for direct quantification of the collision probability. Using this property, they formulate a chance constraint trajectory optimization.

\begin{table*}[!htbp]
\centering
\footnotesize
\setlength{\tabcolsep}{3pt}
\caption{Comparison of our work with related works on Gaussian Splats}
\label{tab:comparison}
\begin{tabular}{|l|c|c|c|c|c|c|}
\hline
& \textbf{Splat-Nav \cite{Chen2024}} & \textbf{GS-Planner \cite{jin2024gs}} & \textbf{GSM \cite{goel2025distance}} & \textbf{GaussNav \cite{lei2025gaussnav}} & \added{\textbf{ReMoSPLAT \cite{marticorena2025remosplat}}} & \textbf{FOCI (ours)} \\
\hline
Planning pipeline included & \textcolor{green}{\cmark} & \textcolor{green}{\cmark} & \textcolor{red}{\xmark} & \textcolor{green}{\cmark} & \textcolor{red}{\xmark} & \textcolor{green}{\cmark} \\
\hline
Orientation-aware collision & \textcolor{red}{\xmark} & \textcolor{red}{\xmark} & \textcolor{red}{\xmark} & \textcolor{red}{\xmark} & \textcolor{red}{\xmark} & \textcolor{green}{\cmark} \\
\hline
Covariance information leveraged & \textcolor{green}{\cmark} & \textcolor{green}{\cmark} & \textcolor{green}{\cmark} & \textcolor{red}{\xmark} & \textcolor{green}{\cmark} & \textcolor{green}{\cmark} \\
\hline
\end{tabular}
\vspace*{-0.0cm}
\end{table*}

Due to its novelty, few papers have been published on trajectory optimization for Gaussian Splats. In GaussNav~\cite{lei2025gaussnav}, Lei et al. reduce the Gaussian Splat to a point cloud of its means, which is voxelized and then projected to a two-dimensional cost map where trajectories are planned. This produces quick results but does not fully leverage the explicit 3D nature of \ac{3dgs}.
Splat-Nav~\cite{Chen2024} additionally takes the covariances of the Gaussians into account and computes a polyhedral safety corridor around an A* initial guess. The free space is determined by calculating the distance between the robot ellipsoid and the confidence intervals of the environment Gaussians. Inside this safety corridor, a spline trajectory is optimized to produce a safe path. 
The GS-Planner~\cite{jin2024gs} constrains the position along the trajectory to be outside of the $3\sigma$ confidence interval of the joint Gaussian of the ellipsoid and robot radius. Unlike Splat-Nav~\cite{Chen2024}, which uses a safety corridor as an intermediate representation, this constraint is directly imposed on the spline.
Goel and Tabib~\cite{goel2025distance} propose a method to derive an \ac{esdf} and collision probability of a Gaussian Surface Model (GSM) for an ellipsoidal robot with a fixed orientation, which can then be used to plan trajectories. \added{Finally, ReMoSPLAT~\cite{marticorena2025remosplat} introduces a reactive controller that solves a quadratic program for collision avoidance, representing the robot as a sphere set and computing distances either directly to the ellipsoids or through multiple virtual-camera views.}

While these works present considerable steps towards \ac{3dgs} planning, they consider only ellipsoidal \added{or spherical} robots with a fixed orientation.
However, in real-world settings~\cite{dong2024efficient, Funk2023}, optimizing orientation is crucial to navigating through narrow passageways.
\textbf{By leveraging a detailed whole-body overlap model, we propose a solution to this shortcoming}.
Table~\ref{tab:comparison} provides an overview of how our algorithm compares to related work.

\section{Method}

Our methodology can be split into three parts:
\textit{1)} trajectory representation to create an initial spline, \textit{2)} collision measure and \textit{3)} optimization loop. \added{In the following we will use the definitions shown in Table \ref{tab:notation}}. 

\begin{table}
\renewcommand{\arraystretch}{1.1}
\footnotesize
\setlength{\tabcolsep}{2.8pt}
\centering
\caption{\added{Summary of the notation used in this work.}}
\label{tab:notation}
\added{
\begin{tabularx}{\columnwidth}{|>{\raggedright\arraybackslash}p{0.35\columnwidth}|X|}
\hline
\textbf{Symbol} & \textbf{Meaning} \\ \hline
$\boldsymbol{b}\in \mathbb{R}^{J_v}$, $\boldsymbol{\theta}\in \mathbb{R}^{J_r}$ & Configuration of the virtual ($\boldsymbol{b}$, $J_v$ joints) and real ($\boldsymbol{\theta}$, $J_r$ joints) joints of the robot.\\ \hline 
$J := J_v + J_r$, $\mathcal{J} := \{1,...,J\}$ & Total number of joints of the robot, and its index set. \\ \hline
\rule{0pt}{2.3ex}$\mathcal{\bar{J}}$ & Set over which spherical velocity and acceleration joint constraints are imposed. Corresponds to the indices of the virtual joints $\boldsymbol{b}$. $\mathcal{\bar{J}}\subseteq \mathcal{J}$\\ \hline
$\boldsymbol{q} := 
\begin{bmatrix}
\boldsymbol{b}^T & \boldsymbol\theta^T
\end{bmatrix}^T \in \mathbb{R}^{J}$ & Configuration of all the joints of the robot. \\ \hline
$N, M \in \mathbb{N}$ & Number of Gaussians that define the environment ($N$) and the robot ($M$). \\ \hline
$K, H \in \mathbb{N}$ & Number of discretization points along the trajectory ($K$) and of spline control points ($H$). \\ \hline
$\boldsymbol{c}_h \in \mathbb{R}^{J}$, $h = 1,\dots,H$ & Control points of the spline trajectory, stacked in $\boldsymbol{C} = [\boldsymbol{c}_1\ \dots\ \boldsymbol{c}_H]^T \in \mathbb{R}^{H \times J}$. \\ \hline
$s \in \mathbb{R}$ & Progress variable of the spline \\ \hline
$\boldsymbol{q}(t) \in \mathbb{R}^{J}$ & Joint angles trajectory evaluated at time $t$. Note that $\boldsymbol{q}$ consists of $J$ splines, one per joint. \\ \hline
$\boldsymbol{q}_j(t) \in \mathbb{R}$ & $j$-th component of $\boldsymbol{q}(t)$. \\ \hline
$\boldsymbol{q}_{\text{start}} \in \mathbb{R}^{J}$ & Joint angles of the robot at the start of the optimization (chosen by user). \\ \hline
$\boldsymbol{f}_{\text{task}}(\boldsymbol{q}(t)) \in \mathbb{R}^3$ & Forward kinematics function that maps the joint angles $\boldsymbol{q}(t)$ at time $t$ into a Cartesian position in the world frame (e.g., end-effector or body position depending on the robot).  \\ \hline
$\boldsymbol{\mu}_n(t) \in \mathbb{R}^3$, $\boldsymbol{\Sigma}_n(t) \in \mathbb{R}^{3 \times 3}$ & Mean and covariance (expressed in the world frame) of environment Gaussian $n$ at time $t$ (constant for static environments). \\ \hline
$\boldsymbol{\bar{\mu}}_{m}(\boldsymbol{q}(t)) \in \mathbb{R}^3$, $\boldsymbol{\bar{\Sigma}}_m(\boldsymbol{q}(t)) \in \mathbb{R}^{3 \times 3}$ & Mean and covariance (expressed in the world frame) of user-defined robot Gaussian $m$ for the joint angles $\boldsymbol{q}(t)$ at time $t$. \\ \hline
$\boldsymbol{\Sigma}_{mn}(t) := \boldsymbol{\Sigma}_n(t) + \bar{\boldsymbol{\Sigma}}_m(\boldsymbol{q}(t)) \in \mathbb{R}^{3 \times 3}$ & Joint covariance of environment Gaussian $n$ and robot Gaussian $m$. \\ \hline
$v_{\text{max}}, a_{\text{max}} \in \mathbb{R}_{>0}$ & 
Maximum absolute virtual joints speed and acceleration, used to enforce kinematic constraints. \\ \hline
$\omega_{j,\text{max}}, \alpha_{j,\text{max}} \in \mathbb{R}_{>0}$ & Maximum absolute real joint speed and acceleration of joint $j$, used to enforce kinematic constraints. \\ \hline
$T\in \mathbb{R}_{>0}$ & 
Total time of the trajectory \\ \hline
$t_k\in \mathbb{R}_{\ge0}$, $k = 1,\dots,K$ & Equidistant discretization times \\ \hline
$\boldsymbol{p}_{\text{goal}}\in \mathbb{R}^3$ & 
Desired goal position for $\boldsymbol{f}_{\text{task}}(\boldsymbol{q}(T))$, expressed in the world frame (e.g., end-effector position for a manipulator, or body position for legged and aerial robots). \\ \hline
\end{tabularx}}
\vspace*{-0.5cm}
\end{table}

\subsection{Trajectory Representation}
We employ cubic B-splines for the optimization to generate fully differentiable and smooth trajectories in $d$ dimensions.
With B-splines, every point along the trajectory can be expressed as a weighted sum of the $H$ control points \added{$\boldsymbol{c}_i$} of the trajectory. Since we are using unclamped uniform cubic B-splines, each spline segment is parameterized by $4$ control points, resulting in $H - 3$ individual segments.
The progress along the trajectory is usually parameterized by a progress variable $s$, which can (but is not required to) be equal to the system time $t$. 
Using knots $\{-3, -2, \dots, H -1, H\}$, the spline is parameterized on $s \in [0, H - 3)$. Since each individual spline segment is parameterized between $[0,1]$, we define $s' = s - i$ with $i :=\lfloor s \rfloor$ to index the corresponding segment. For segment $i$, the spline can be evaluated with
\added{
\begin{equation}
\label{eq:point_wise_spline}
\setlength{\arraycolsep}{3pt} %
    \boldsymbol{q}^T(s') = 
\begin{bmatrix}
1 & {s'} & {s'}^2 & {s'}^3
\end{bmatrix}
\frac{1}{6}
\begin{bmatrix}
1 & 4 & 1 & 0 \\
-3 & 0 & 3 & 0 \\
3 & -6 & 3 & 0 \\
-1 & 3 & -3 & 1
\end{bmatrix}
\begin{bmatrix}
\boldsymbol{c}_{i+1}^T \\
\boldsymbol{c}_{i+2}^T \\
\boldsymbol{c}_{i +3}^T \\
\boldsymbol{c}_{i +4}^T
\end{bmatrix}
\end{equation}
}

\added{We evaluate the spline at $K$ equidistant times $t_k = (k-1)T/(K-1)$, $k = 1,\dots,K$, and stack the control points in $\boldsymbol{C} = [\boldsymbol{c}_1\ \dots\ \boldsymbol{c}_H]^T \in \mathbb{R}^{H \times d}$. Time derivatives follow from $\frac{d^i \boldsymbol{q}}{dt^i} = \frac{d^i \boldsymbol{q}}{ds^i}\left(\frac{ds}{dt}\right)^i$, where $\frac{ds}{dt}$ is assumed constant.} This parameterization works well for simple vector spaces such as $\mathbb{R}^3$ but fails to interpolate between orientations. 
Instead of the full pose, \textbf{only the joint angles $\boldsymbol{q} \in \mathbb{R}^J$ are optimized}. In the following, we assume $d = J$ and decompose the trajectory evaluations as $\boldsymbol{q} = [\boldsymbol{b}^T, \boldsymbol{\theta}^T]^T \in \mathbb{R}^J$. \added{The \emph{virtual} joints $\boldsymbol{b}$ are not physical joints: they can be seen as fictitious prismatic joints that connect the world frame to the base of the robot and describe its translation. The \emph{real} joints $\boldsymbol{\theta}$ are the remaining degrees of freedom. For example, for a mobile-base manipulator with a six-joint arm moving on the ground plane, $\boldsymbol{b} = [x, y]^T$ is the position of the base ($J_v = 2$) and $\boldsymbol{\theta} = [\text{yaw}, \theta_1, \dots, \theta_6]^T$ stacks the yaw of the base and the arm joint angles ($J_r = 7$), whereas a fixed-base manipulator has no virtual joints ($J_v = 0$).}

\subsection{Collision Measure}
To quantify the collision between two objects, we want to measure the overlap between them. The collision measure should, therefore, fulfill the following three properties:
\textit{1)} It should be close to zero if two objects do not overlap.
\textit{2)} It should be greater than zero when two objects overlap and quantify the magnitude of that overlap.
\textit{3)} It should be differentiable to allow for gradient evaluations in optimization.

\added{In our work, we model both the environment $p(\boldsymbol{x}, t)$ and the robot $r(\boldsymbol{x}, \boldsymbol{q}(t))$ by 3D Gaussian Splats. The environment is formed by $N$ Gaussians, and the robot by $M$ Gaussians placed at user-defined points. This placement differs for each robotic platform. In the case of legged and aerial systems, the Gaussians are placed according to an offset relative to the center of the robot, while for robotic manipulators each Gaussian is associated with a link. Note that not all links of the robot need to have Gaussians associated with them, since some links may act as dummy links. We define the density of the robot and the environment at a specific point  $\boldsymbol{x} \in \mathbb{R}^3$  in space and time $t \in \mathbb{R}$ as}
\added{\begin{equation*}
    p(\boldsymbol{x}, t) = \sum_{n=1}^N\mathcal{N}(\boldsymbol{x};\bm{\mu}_n(t),\boldsymbol{\Sigma}_n(t)) \in \mathbb{R}
\end{equation*}}
\added{\begin{equation*}
    r(\boldsymbol{x}, \boldsymbol{q}(t)) = \sum_{m=1}^M\mathcal{N}(\boldsymbol{x};\bm{\bar{\mu}}_m(\boldsymbol{q}(t)),\bar{\boldsymbol{\Sigma}}_m(\boldsymbol{q}(t))) \in \mathbb{R}
\end{equation*}}

\added{We define the collision volume as the region where both quantities overlap. To quantify the collision magnitude, we compute the overlap integral between the environment field $p(\boldsymbol{x}, t)$ and the robot field $r(\boldsymbol{x}, \boldsymbol{q}(t))$:}

\added{\footnotesize\begin{gather}
\label{eq:convolution}
\int_{\mathbb{R}^3} p(\boldsymbol{x}, t) r(\boldsymbol{x}, \boldsymbol{q}(t)) d\boldsymbol{x} \\
 =\sum_{n=1}^N \sum_{m=1}^M 
\int_{\mathbb{R}^3}
\mathcal{N}(\boldsymbol{x};\bm{\mu}_n(t),\boldsymbol{\Sigma}_n(t))
\mathcal{N}(\boldsymbol{x};\bm{\bar{\mu}}_m(\boldsymbol{q}(t)),\bar{\boldsymbol{\Sigma}}_m(\boldsymbol{q}(t)))
\, d\boldsymbol{x} \nonumber \\
= \sum_{n=1}^N \sum_{m=1}^M 
\mathcal{N}(\bm{\bar{\mu}}_m(\boldsymbol{q}(t));\bm{\mu}_n(t),\boldsymbol{\Sigma}_n(t) + \bar{\boldsymbol{\Sigma}}_m(\boldsymbol{q}(t))) \nonumber
\end{gather}}

\added{This results in a representation that is fully differentiable with respect to $\boldsymbol{q}$. Note that the robot means and covariances are parameterized by the robot joint angles -- whose gradients can be derived with the chain rule -- while the environment means and covariances depend on the time $t$, allowing dynamic scenarios.}

This measure for collision fulfills the desired properties defined above.
Because of the Gaussian's exponential nature, it converges to zero as the distance between the robot and the obstacle increases. That also means it does not contribute meaningfully to the gradient at farther distances.

Note that \added{each term in Equation~\ref{eq:convolution} is a scaled} exponential of \added{minus half the squared} Mahalanobis distance between \added{the robot and environment Gaussians, i.e., $\mathcal{N}(\bm{\bar{\mu}}_m;\bm{\mu}_n,\boldsymbol{\Sigma}_{mn}) = \det(2\pi\boldsymbol{\Sigma}_{mn})^{-1/2} \exp(-\tfrac{1}{2}\bm{\delta}_{mn}^T\boldsymbol{\Sigma}_{mn}^{-1}\bm{\delta}_{mn})$, with $\bm{\delta}_{mn}(t) = \bm{\bar{\mu}}_m(\boldsymbol{q}(t)) - \bm{\mu}_n(t)$ and the joint covariance $\boldsymbol{\Sigma}_{mn}(t) := \boldsymbol{\Sigma}_n(t) + \bar{\boldsymbol{\Sigma}}_m(\boldsymbol{q}(t))$}.
The Mahalanobis distance \added{accounts for} the potentially \added{anisotropic} extent of the Gaussians, and the exponential ensures the desired \added{decay}.

\subsection{Optimization Formulation}

Whenever integrating a quantity like the collision measure along the spline is necessary, we approximate it as a discrete sum over the values discretized along the spline in $K$ equidistant steps.
\added{We represent the robot joints as cubic B-splines. The position of each Gaussian comprising the robot is a function of both virtual and real joint angles $\bar{\bm{\mu}}_m(\boldsymbol{b}(t), \boldsymbol{\theta}(t)) = \bar{\bm{\mu}}_m(\boldsymbol{q}(t))$.}

\added{We minimize the weighted sum of three terms: the obstacle cost, the jerk cost, and the goal (distance-to-target) cost. The jerk cost is divided for real and virtual joints. }

\added{The initial joint angles are constrained to $\boldsymbol{q}_{\text{start}}$, and the goal is imposed by minimizing the distance between the final value of the task-space function $\boldsymbol{f}_{\text{task}}$ and the goal position. Hence, the meaning of the goal depends on the platform: for a manipulator, $\boldsymbol{f}_{\text{task}}$ is the end-effector position (obtained via forward kinematics), while for a UAV and a legged robot it is the center of the robot body.}

Instead of constraining samples of the \added{spline derivatives}, we leverage the convex hull property of splines.
Since each spline segment lies within the convex hull of its control points, it is enough to constrain the norm of the velocity and acceleration control points to guarantee constraint satisfaction along the entire spline. \added{To minimize} the conservativeness of this overapproximation, we compute the convex hull with respect to the \textbf{MINVO Basis}~\cite{tordesillas2022minvo}, \added{which yields the minimum-volume simplex enclosing each spline segment}.

\added{Each real joint $j \in \mathcal{J} \setminus \mathcal{\bar{J}}$ has angular velocity and acceleration bounds $\omega_{j,\text{max}}$ and $\alpha_{j,\text{max}}$, while the virtual joints $j \in \mathcal{\bar{J}}$ are bounded jointly in norm by $v_{\text{max}}$ and $a_{\text{max}}$, which prevents faster motion along diagonal directions than along individual axes. \added{The virtual and real joints of each robot are summarized in Table~\ref{tab:joints}.}}

The optimization problem is defined in Equation \ref{eq:cost}.
\added{\small
\begin{align}
\min_{\boldsymbol{C}} \enspace 
& w_1 \cdot \frac{1}{KNM} \sum_{k=1}^{K}\sum_{n=1}^{N}\sum_{m=1}^M  \mathcal{N}(\boldsymbol{\bar{\mu}}_m(\boldsymbol{q}(t_k));\boldsymbol{\mu}_n(t_k), \boldsymbol{{\Sigma}}_{mn}(t_k)) \nonumber \\
& + w_{2} \cdot \frac{T}{(K-1) \cdot |\mathcal{J} \setminus \bar{\mathcal{J}}|} \sum_{k=1}^{K} \sum_{j\in\mathcal{J} \setminus \bar{\mathcal{J}}} \left(\boldsymbol{\dddot{q}}_j(t_k)\right)^2  \nonumber \\& + w_{3}\cdot \frac{T}{(K-1) \cdot \lvert \mathcal{\bar{J}} \rvert}\sum_{k=1}^{K}\sum_{j\in\mathcal{\bar{J}}} \left(\boldsymbol{\dddot{q}}_j(t_k)\right)^2  \nonumber \\
& + w_4 \left\lVert \boldsymbol{f}_{\text{task}}(\boldsymbol{q}(T)) - \boldsymbol{p}_{\text{goal}} \right\rVert^2 \nonumber \\
\text{s.t.} \enspace 
& \boldsymbol{q}(0) = \boldsymbol{q}_{\text{start}} \nonumber \\
& \left\lVert [\dot{\boldsymbol{q}}_j(t)]_{j\in \bar{\mathcal{J}}} \right\rVert^2 \leq v_{\text{max}}^2, \quad \forall t \in [0, T] \nonumber \\
& \left\lVert [\ddot{\boldsymbol{q}}_j(t)]_{j\in \bar{\mathcal{J}}} \right\rVert^2 \leq a_{\text{max}}^2, \quad \forall t \in [0, T] \nonumber \\
& \left| \dot{\boldsymbol{q}}_j(t) \right| \leq \omega_{j,\text{max}}, \quad \forall j \in \mathcal{J} \setminus \bar{\mathcal{J}}, \ \forall t \in [0, T] \nonumber \\
& \left| \ddot{\boldsymbol{q}}_j(t) \right| \leq \alpha_{j,\text{max}}, \quad \forall j \in \mathcal{J} \setminus \bar{\mathcal{J}}, \ \forall t \in [0, T] \label{eq:cost}
\end{align}}
\added{Terms and constraints with an empty index set are omitted (e.g., $\bar{\mathcal{J}} = \emptyset$ for a fixed-base manipulator).}

Due to the non-linear and constrained nature of the optimization problem, we employ an interior point method.
In order to create an initial guess, the \ac{3dgs} is used to create a rough occupancy grid, where a voxel is occupied if it contains at least one mean. \added{For mobile robots and UAVs, A* or RRT* can be run on this occupancy grid to obtain an initial path, which is then used to initialize the control points by fitting a spline. For robotic manipulators  where planning is naturally posed in high-dimensional configuration spaces, only RRT* is used for initialization.}

\subsection{Implementation Details}
\label{sec:implementation}
We define and solve the described optimization problem with CasADi~\cite{andersson2019casadi}. We exploit the independent summation structure of the collision measure (Equation \ref{eq:convolution}) and run this computation on the GPU in parallel. This led us to define a custom CasADi functor that computes the evaluation of the function and its gradients in parallel using NVIDIA Warp~\cite{warp2022}.

\added{We consider two variants of the proposed robot representation. The first employs spherical covariance matrices, making the robot covariances independent of the robot configuration. For static environments, where obstacle covariances are also constant, the covariance sums in Equation~\ref{eq:convolution} can thus be precomputed, reducing the computational cost of trajectory optimization. The second variant uses anisotropic covariance matrices; this is useful, for instance, to better approximate elongated robot links, yielding a more accurate robot representation at the expense of additional computation.}

Note that other GPU extensions for CasADi, such as L4Casadi~\cite{salzmann2023l4casadi} and CusADI~\cite{jeon2024cusadigpuparallelizationframework} exist, but were not directly applicable due to the custom 3D Gaussian overlap integral formulation.
The optimization problem is then solved via the interior point method (IPOPT)~\cite{wachter2006implementation} with the custom overlap integral functor \added{using the linear solver MA27~\cite{duff1982ma27, hsl}.}

\section{Experiments}
In this section, we evaluate our algorithm by applying it to planning problems of different robots in different \ac{3dgs} environments, both artificially generated scenes and ones created from real-world captures.\footnote{See the Readme of the released code for the links and attribution of each \ac{3dgs}  and robot used} \added{Table~\ref{tab:joints} lists the virtual and real joints of each robot used in the experiments. For the quadruped, only the trajectory of the base is planned, and the leg motions are generated by a lower-level locomotion policy~\cite{locomotion2022}. Both UAVs share the same joints but differ in their dynamics: the standard (underactuated) UAV's differential flatness couples the orientation and the acceleration while the  omnidirectional UAV does not have this coupling.} All \added{ANYmal} experiments are run on an Intel i7-8750H with 16 GB of RAM and an NVIDIA RTX 2070 Max-Q\added{, and the rest of the experiments are run on an Intel i7-13700 with 64 GB of RAM and an NVIDIA GeForce RTX 4070 Ti}.

\begin{table}[!h]
\centering
\scriptsize
\setlength{\tabcolsep}{3pt}
\caption{\added{Virtual ($\boldsymbol{b}$) and real ($\boldsymbol{\theta}$) joints of each robot.}}
\label{tab:joints}
\added{%
\begin{tabular}{lcccc}
\toprule
\textbf{Robot} & $\bm{J_v}$ & \textbf{Virtual joints} $\boldsymbol{b}$ & $\bm{J_r}$ & \textbf{Real joints} $\boldsymbol{\theta}$ \\
\midrule
ANYmal (quadruped) & 2 & $[x, y]^T$ & 1 & yaw \\
Manipulator (UR5) & 0 & -- & 6 & arm joint angles $\theta_1, \dots, \theta_6$ \\
Omnidir. UAV & 3 & $[x, y, z]^T$ & 1 & yaw \\
Standard UAV & 3 & $[x, y, z]^T$ & 1 & yaw \\
\bottomrule
\end{tabular}}
\vspace{-0.3cm}
\end{table}

\added{For FOCI and Splat-Nav, a warm-up run is performed first and each planning query is then executed three times; the reported time is the median of the three runs, so that it does not depend on the state of the GPU. CHOMP and STOMP are run once. As a baseline (Table~\ref{tab:stonehenge}), RRT* is run once with a fixed planning time budget of $100\,\si{\s}$; as an initializer (Table~\ref{tab:results}), it stops once it reaches the goal or stalls (at most $10\,\si{\s}$). The path length is reported as a dimensionless quantity, normalized by the diagonal of the scene.}

\subsection{ANYmal Experiments}
\subsubsection{Simulation}
\begin{figure}[tb]
    \centering
    \begin{subfigure}[b]{0.49\linewidth}
        \centering
        \includegraphics[width=\linewidth]{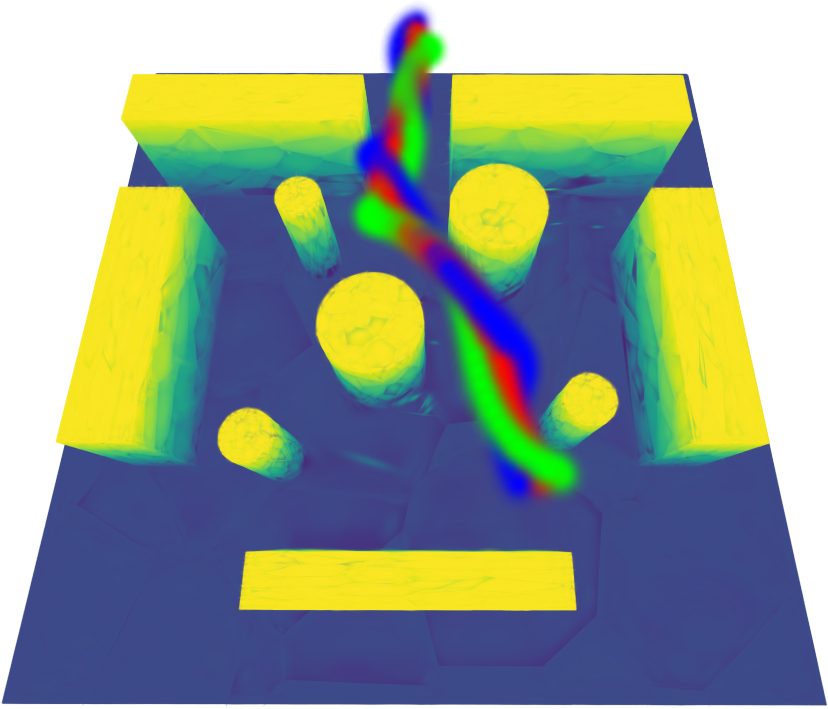}
        \caption{Pillar Environment}
        \label{fig:complex_small}
    \end{subfigure}
    \hfill
    \begin{subfigure}[b]{0.49\linewidth}
        \centering
        \includegraphics[width=\linewidth]{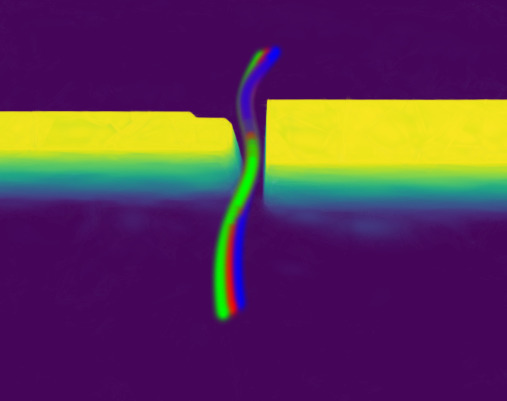}
        \caption{Narrow Corridor}
        \label{fig:narrow-corridor}
    \end{subfigure}
    \caption{Sample trajectories created on synthetic testing data showing a 3 Gaussian robot rotating to navigate the environments\added{ (green: front, red: middle, blue: rear)}.}
    \label{fig:ANYmal_splats}
     \vspace{-0.0cm}
\end{figure}

\begin{figure}[tb]
    \centering
    \begin{subfigure}[b]{0.49\linewidth}
        \centering
        \includegraphics[width=\linewidth]{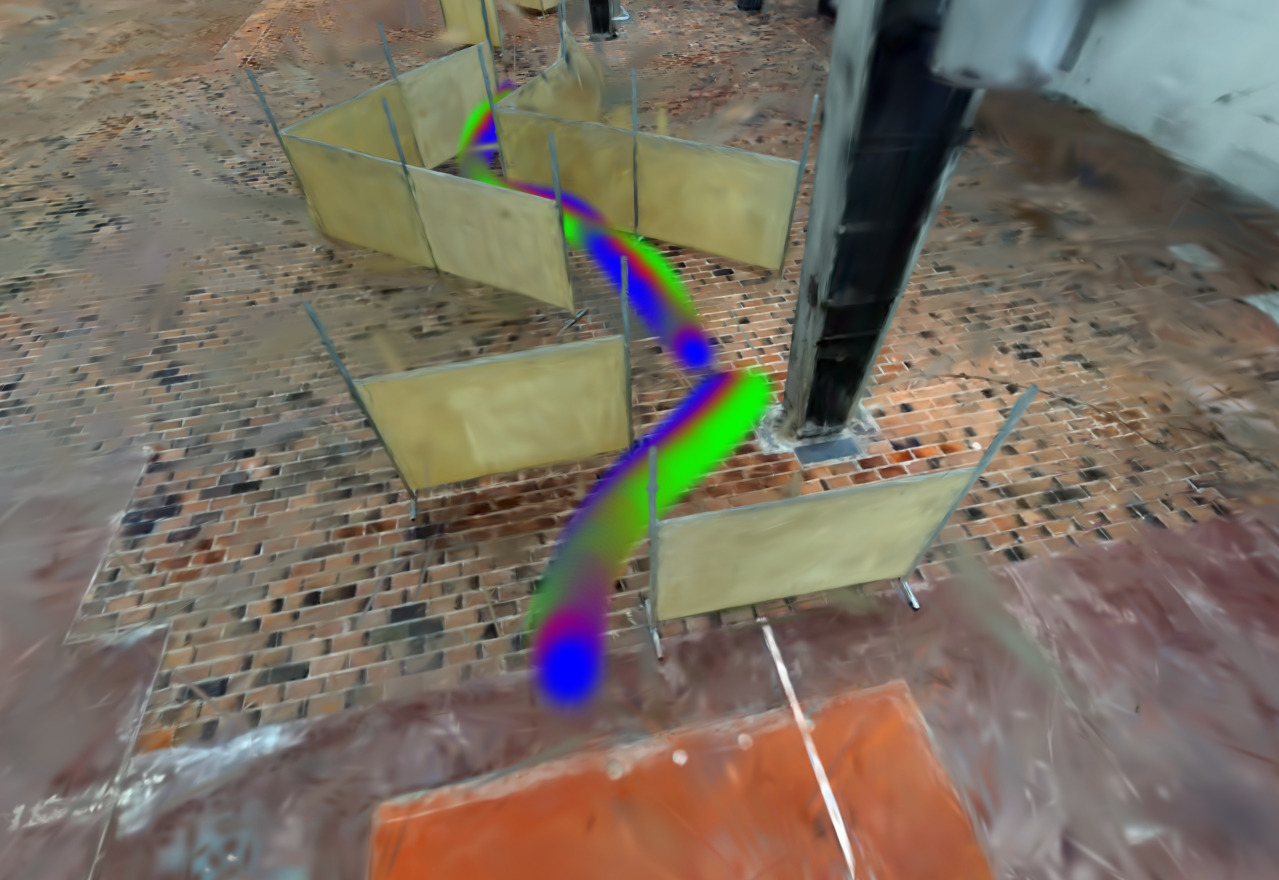}
        \caption{Machine Hall}
        \label{fig:conference_room}
    \end{subfigure}
    \hfill
    \begin{subfigure}[b]{0.49\linewidth}
        \centering
        \includegraphics[width=\linewidth]{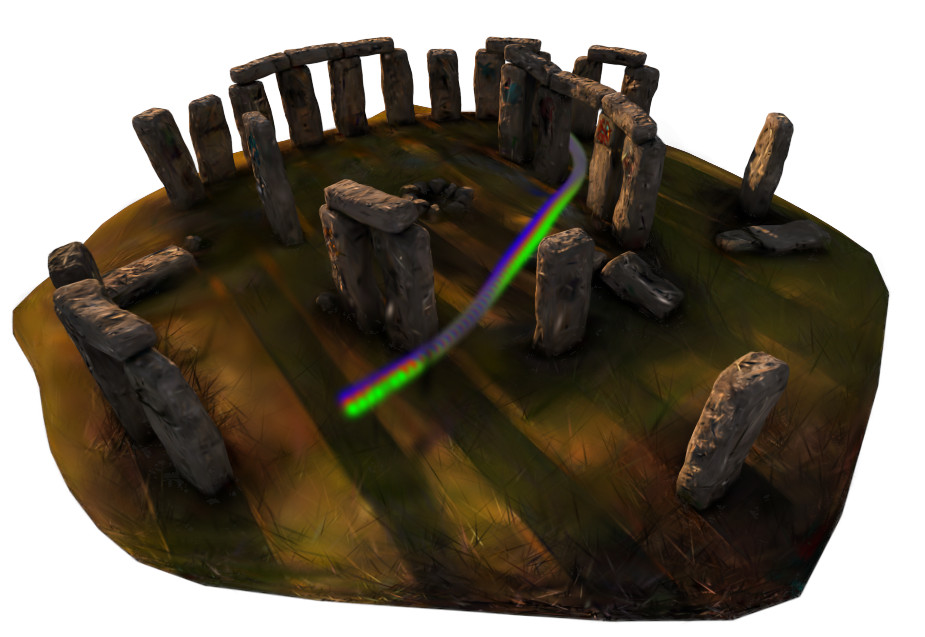}
        \caption{Stonehenge Environment}
        \label{fig:rsl}
    \end{subfigure}
    \caption{Trajectories planned for ANYmal through realistic scenes.}
    \label{fig:real_splats}
     \vspace*{-0.0cm}
\end{figure}

 \added{The results obtained for the ANYmal robot  (modeled as a 3-Gaussian robot) in both synthetic and realistic scenes are reported in Table~\ref{tab:anymal} and Fig.~\ref{fig:ANYmal_splats}. The solve time stays below $0.5\,\si{\s}$ in the synthetic scenes and increases in the larger realistic scenes, up to $2.12\,\si{\s}$ in Machine Hall (243k Gaussians), while the initial guess takes at most $0.55\,\si{\s}$ in all scenes. Fig.~\ref{fig:complex_small} shows longer trajectories through the more complex pillar environment. The algorithm gracefully handles trajectories requiring additional turns while adapting the orientation to keep a maximum distance from the wall.  
 	Fig.~\ref{fig:narrow-corridor} shows how FOCI effectively leverages the asymmetry of ANYmal to pass through the narrow opening collision-free. The start and end poses of the robot are constrained to be parallel to the wall, requiring the orientation to change to pass through the opening.
 	Finally, Fig.~\ref{fig:real_splats} shows that we can plan collision-free trajectories through more realistic \ac{3dgs}. In the case of the machine hall sample, the data used for the reconstruction came entirely from the onboard sensors of the robot~\cite{patil2024radiance}, quickly providing a scene with accurate scaling.
 }

\begin{table}[t]
\centering
\scriptsize
\setlength{\tabcolsep}{4pt}
\caption{\added{Planning time of FOCI for the ANYmal robot in synthetic (Narrow Corridor, Pillars) and realistic (Machine Hall, Stonehenge) \ac{3dgs} scenes. $N$/$M$: number of environment/robot Gaussians.}}
\label{tab:anymal}
\begin{tabular}{clcccc}
\toprule
\added{\textbf{Robot}} & \textbf{Environment} & \textbf{N} & \textbf{M} & \textbf{\begin{tabular}[c]{@{}c@{}}Init. Guess \\ Time (s)\end{tabular}} & \textbf{\begin{tabular}[c]{@{}c@{}}Solve\\ Time (s)\end{tabular}} \\
\midrule
\multirow{4}{*}{\rotatebox{90}{ANYmal}} & Narrow Corridor & 24k & 3 & $0.24$ & $0.47$ \\
 & Pillars & 49k & 3 & $0.25$ & $0.45$ \\
\cmidrule(l){2-6}
 & Machine Hall & 243k & 3 & $0.22$ & $2.12$ \\
 & Stonehenge & 138k & 3 & $0.55$ & $0.83$ \\
\bottomrule
\end{tabular}
 \vspace{-0.0cm}
\end{table}

\subsubsection{Hardware}

\begin{figure}[t]
	\centering
	\includegraphics[width=0.92\linewidth]{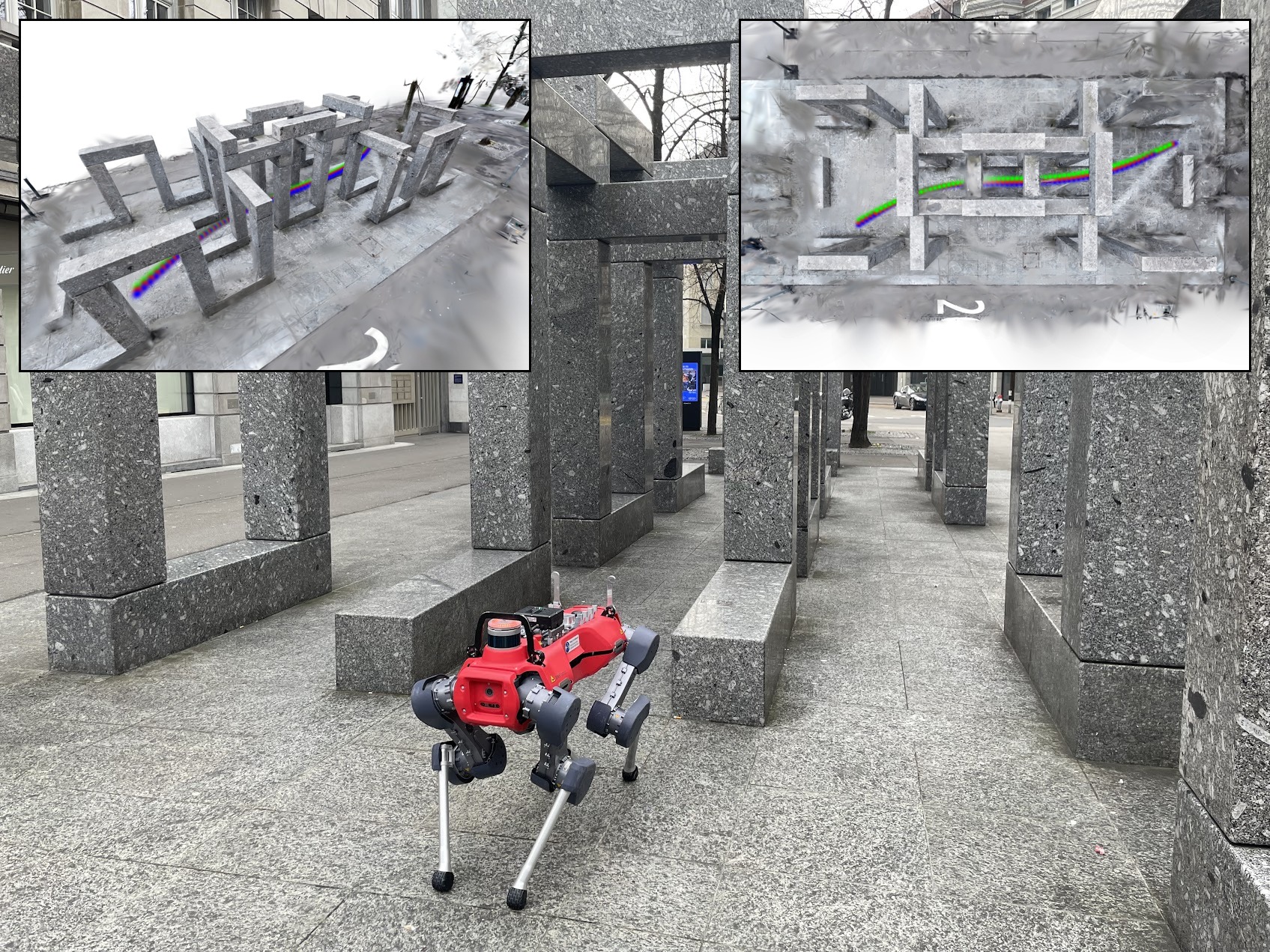}
	\caption{Trajectories planned for the ANYmal robot through real-world obstacles. The actual robot can be seen at the site following the trajectories generated by our algorithm \textbf{FOCI}.}
	\label{fig:realworld}
	\vspace{-0.0cm}
\end{figure}

We also showcase FOCI deployed on hardware using a real \ac{3dgs} reconstruction of a scene captured using UAV imagery and localized on site using ICP on the LiDAR data and Gaussian means. The narrow passages visible in Fig.~\ref{fig:realworld} required the robot to rotate in order to safely traverse the environment, while obstacles on the ground forced the height-constrained trajectories to avoid shortcuts.

\subsection{\added{Manipulator Experiments}}

\added{Manipulator trajectory optimization is particularly challenging due to the high dimensionality and non-convexity of the configuration space. We evaluate our method with a UR5 manipulator in two \ac{3dgs} environments, Bonsai and Coffee Table. We further compare the proposed approach against CHOMP~\cite{zucker2013chomp} and STOMP~\cite{kalakrishnan2011stomp}. Since STOMP is a local optimizer, its performance is highly affected by the quality of the initial guess. 
For a manipulator, the Cartesian goal corresponds to a lower-dimensional region of the configuration space with no closed-form mapping to guide the search, making it harder for RRT* to find a valid configuration.
Both baselines are our own implementations of~\cite{zucker2013chomp} and~\cite{kalakrishnan2011stomp}, adapted to solve the same problem as FOCI. These implementations are available in the released code. Both CHOMP and STOMP use the same CasADi-based collision functor and are initialized identically to FOCI (see the Table~\ref{tab:results} caption for details). STOMP is marked as not converged if its cost is still improving after $1000$ iterations.}

\begin{figure}[t]
    \centering
    \begin{subfigure}[b]{0.49\linewidth}
        \centering
        \includegraphics[width=\linewidth]{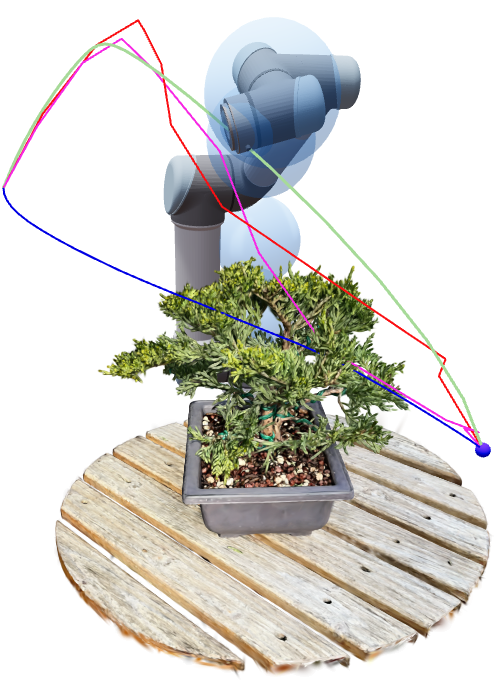}
        \caption{Bonsai environment}
        \label{fig:Bonsai_comparison}
    \end{subfigure}
    \hfill
    \begin{subfigure}[b]{0.49\linewidth}
        \centering
        \includegraphics[width=\linewidth]{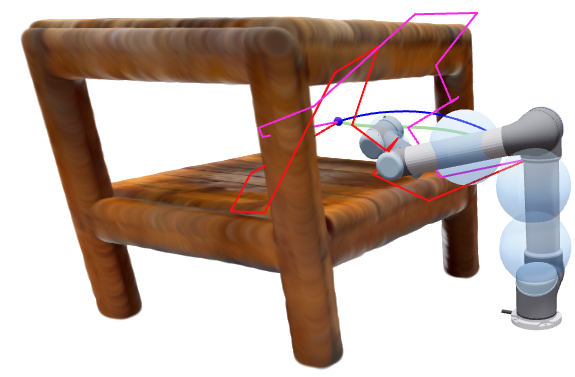}
        \caption{Coffee Table environment}
        \label{fig:Table_comparison}
    \end{subfigure}
    \caption{\added{Qualitative comparison of the trajectories generated by FOCI (green), CHOMP (blue), STOMP (pink) and RRT* (red). The light blue spheres represent the robot Gaussians.}}
    \label{fig:comparison_merged}
    \vspace*{-0.6cm}
\end{figure}

\added{The results are shown in Table~\ref{tab:results} and Fig.~\ref{fig:comparison_merged}. In terms of solve time, FOCI is roughly one order of magnitude faster than CHOMP and two orders of magnitude faster than STOMP in the Bonsai scene; in the Coffee Table scene, this gap widens to roughly two and three orders of magnitude, respectively. STOMP does not converge in either scenario. In the Bonsai scene, CHOMP finds a shorter but colliding path while FOCI remains collision-free; in the Coffee Table scene, FOCI and CHOMP converge to similarly safe trajectories.
}

\begin{table}[t]
\centering
\scriptsize
\setlength{\tabcolsep}{2pt}
\caption{Planning time \added{and normalized path length of FOCI and the baseline methods}. \added{CHOMP and STOMP start from the same initial guess as FOCI (RRT* for the manipulator, a straight line for the UAV).} \added{$N$/$M$: number of environment/robot Gaussians.}
\added{No bold is used in the last column due to the fact that some methods generate short trajectories that collide, see Figs.~\ref{fig:comparison_merged} and \ref{fig:forest}.}
\added{$^\S$The optimizer did not converge.}
}
\label{tab:results}
\begin{tabular}{clcclccc}
\toprule
\added{\textbf{Robot}} & \textbf{Environment} & \textbf{N} & \textbf{M} & \textbf{Method} & \textbf{\begin{tabular}[c]{@{}c@{}}Init. Guess \\ Time (s)\end{tabular}} & \textbf{\begin{tabular}[c]{@{}c@{}}Solve\\Time (s)\end{tabular}} & \textbf{\begin{tabular}[c]{@{}c@{}}\added{Path}\\\added{Length (-)}\end{tabular}} \\
\midrule
\multirow{6}{*}{\rotatebox{90}{\added{Manipulator}}} & \multirow{3}{*}{\added{Bonsai}} & \multirow{3}{*}{\added{$\sim$569k}} & \added{6} & \added{FOCI} & \added{$0.288$} & \added{\textbf{3.059}} & \added{$1.402$} \\
 &  &  & \added{6} & \added{CHOMP} & \added{$0.288$} & \added{$29.276$} & \added{$1.126$} \\
 &  &  & \added{6} & \added{STOMP$^\S$} & \added{$0.288$} & \added{$439.573$} & \added{$1.525$} \\
\cmidrule(l){2-8}
 & \multirow{3}{*}{\added{Coffee Table}} & \multirow{3}{*}{\added{$\sim$668k}} & \added{9} & \added{FOCI} & \added{$4.426$} & \added{\textbf{0.528}} & \added{$0.481$} \\
 &  &  & \added{9} & \added{CHOMP} & \added{$4.426$} & \added{$48.291$} & \added{$0.494$} \\
 &  &  & \added{9} & \added{STOMP$^\S$} & \added{$4.426$} & \added{$582.902$} & \added{$2.035$} \\
\midrule
\multirow{3}{*}{\rotatebox{90}{\added{\begin{tabular}{@{}c@{}}Omnidir.\\UAV\end{tabular}}}} & \multirow{3}{*}{\added{Forest}} & \multirow{3}{*}{\added{$\sim$2.40M}} & \added{1} & \added{FOCI} & \added{$0.0003$} & \added{$0.532$} & \added{$0.169$} \\
 &  &  & \added{1} & \added{CHOMP} & \added{$0.0003$} & \added{\textbf{0.011}} & \added{$0.168$} \\
 &  &  & \added{1} & \added{STOMP} & \added{$0.0003$} & \added{$4.315$} & \added{$0.168$}\\

\bottomrule
\end{tabular}
\end{table}

\subsection{\added{UAV Experiments}}

\subsubsection{Omnidirectional UAV in Static Environment (Forest)}\label{subsubsec:forest}

\begin{figure}[t]
    \centering
    \includegraphics[width=0.79\linewidth]{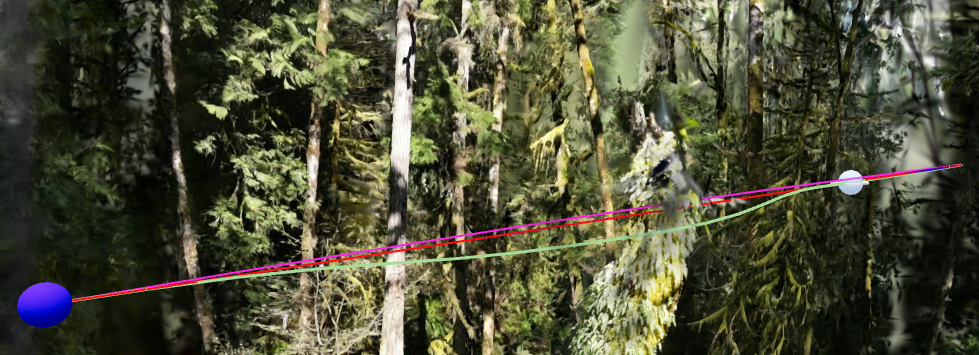}
    \caption{\added{Qualitative comparison of the trajectories generated by FOCI (green), CHOMP (blue), STOMP (pink) and a Straight line initialization (red). The light blue sphere represents the robot Gaussian}}
    \label{fig:forest}
    \vspace{-0.0cm}
\end{figure}

\added{In these experiments, we use an omnidirectional UAV, for which there is no coupling between acceleration and orientation, allowing position and yaw to be optimized independently of the vehicle's tilt. The results are shown in Table~\ref{tab:results} and Fig.~\ref{fig:forest} for a Forest environment, where the straight line between the start and the goal grazes a large tree trunk. As with the manipulator experiments, the proposed approach is compared against CHOMP~\cite{zucker2013chomp} and STOMP~\cite{kalakrishnan2011stomp}; unlike the manipulator scenes, here all methods are initialized with a straight line rather than RRT$^*$, since the large number of Gaussians in this scene ($\sim$2.40M) made RRT$^*$ highly sensitive to its parameters, consistently yielding poor initial guesses for every method. CHOMP and STOMP both converge, but they barely modify the initial guess: CHOMP's trajectory still grazes the trunk, much like the initial guess, while STOMP's trajectory collides with it. In contrast, FOCI moves the trajectory away from the trunk, with a path only $0.7\%$ longer than the straight line.}

\subsubsection{Standard UAV in Dynamic Environment}
\added{To demonstrate the framework's applicability beyond static scenes, Fig.~\ref{fig:mobile_example} shows a standard (i.e., underactuated) UAV planning in an environment with moving obstacles. Unlike the omnidirectional UAV used in Section \ref{subsubsec:forest}, in a standard UAV differential flatness couples the horizontal acceleration to the vehicle's orientation. Because the obstacles' Gaussian parameters are evaluated at each timestep $t$ (Equation~\ref{eq:convolution}), the collision measure -- and consequently the optimized trajectory -- is recomputed as the obstacles move, rather than assuming a static environment. In this experiment, FOCI plans a $13.873\,\mathrm{m}$ trajectory ($1.140$ normalized by the scene diagonal) in $25.318\,\mathrm{s}$ of solve time.} 

\subsection{Runtime}
\begin{figure}[tb]
    \centering
    \begin{subfigure}[b]{0.49\linewidth}
        \centering
        \includegraphics[width=\linewidth]{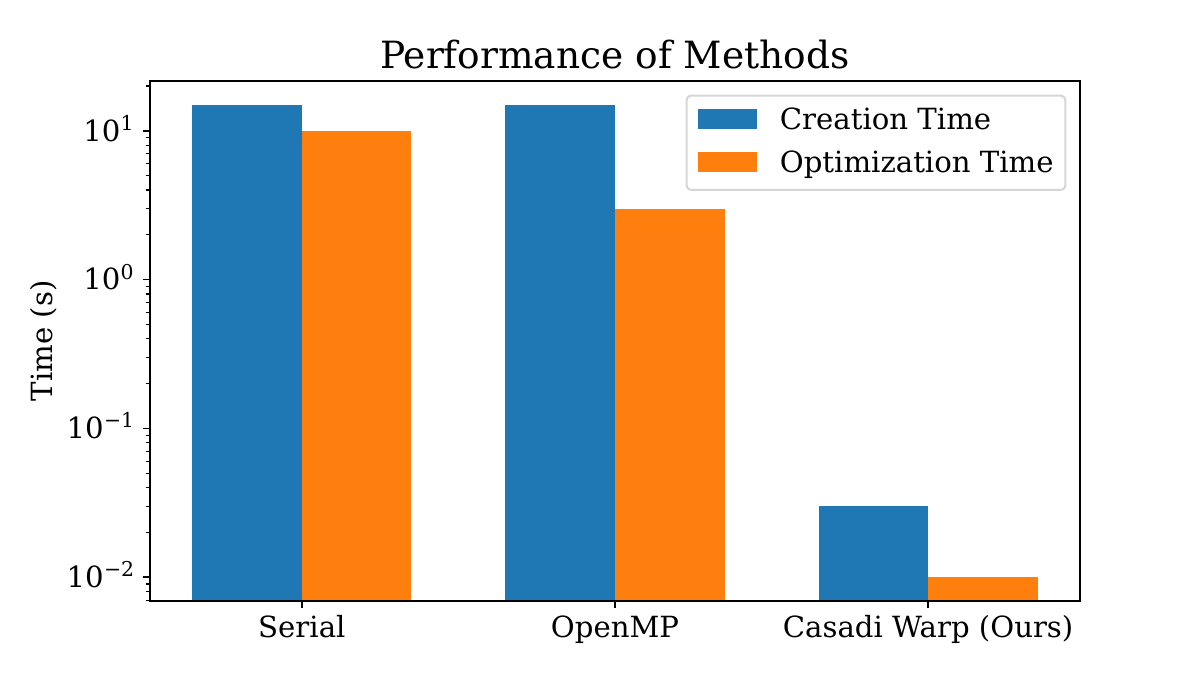}
        \caption{CPU vs.\ GPU runtime}
        \label{fig:runtimes}
    \end{subfigure}
    \hfill
    \begin{subfigure}[b]{0.49\linewidth}
        \centering
        \includegraphics[width=\linewidth]{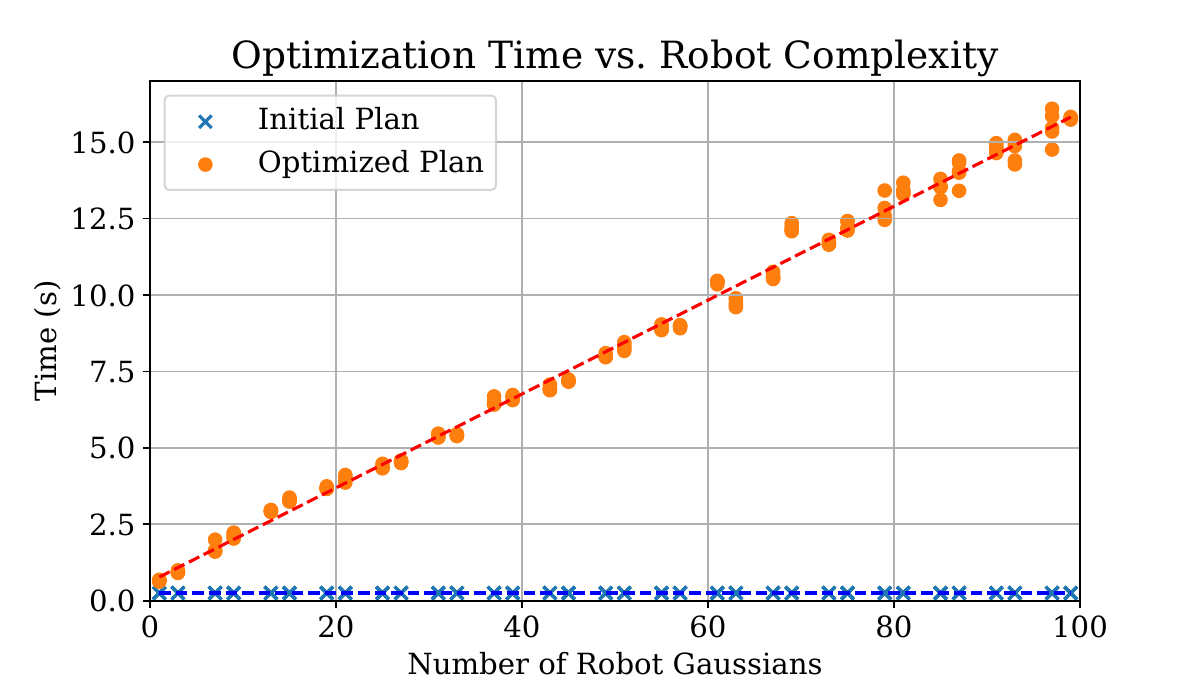}
        \caption{Effect of robot complexity}
        \label{fig:robot_complexity}
    \end{subfigure}
    \caption{(a) Comparison of the solver's creation and runtime on the CPU and GPU for $50$k environmental Gaussians and one robot Gaussian. \added{Serial: single CPU core; OpenMP: multiple CPU cores; CasADi Warp: our GPU implementation.} (b) Optimization time for orientation-aware planning using increasingly complex robot models \added{in Stonehenge (138k Gaussians)}.}
    \label{fig:runtime_complexity_merged}
     \vspace{-0.5cm}
\end{figure}

We now evaluate in Fig.~\ref{fig:runtime_complexity_merged} the performance of our method by comparing the runtimes of the CasADi optimization on a single CPU core, multiple CPU cores, and the GPU.
While a significant speedup can be observed when optimizing on multiple CPU cores using OpenMP~\cite{dagum1998openmp}, {our custom GPU-enabled implementation is $320$ times faster}, often resulting in solutions that only take a few seconds (see Fig.~\ref{fig:runtimes}). Note also the linear relationship between robot complexity and optimization time achieved by FOCI, which shows the potential of modeling complex robot geometry (Fig.~\ref{fig:robot_complexity}).

\begin{figure}[t]
    \centering
    \includegraphics[width=0.92\linewidth]{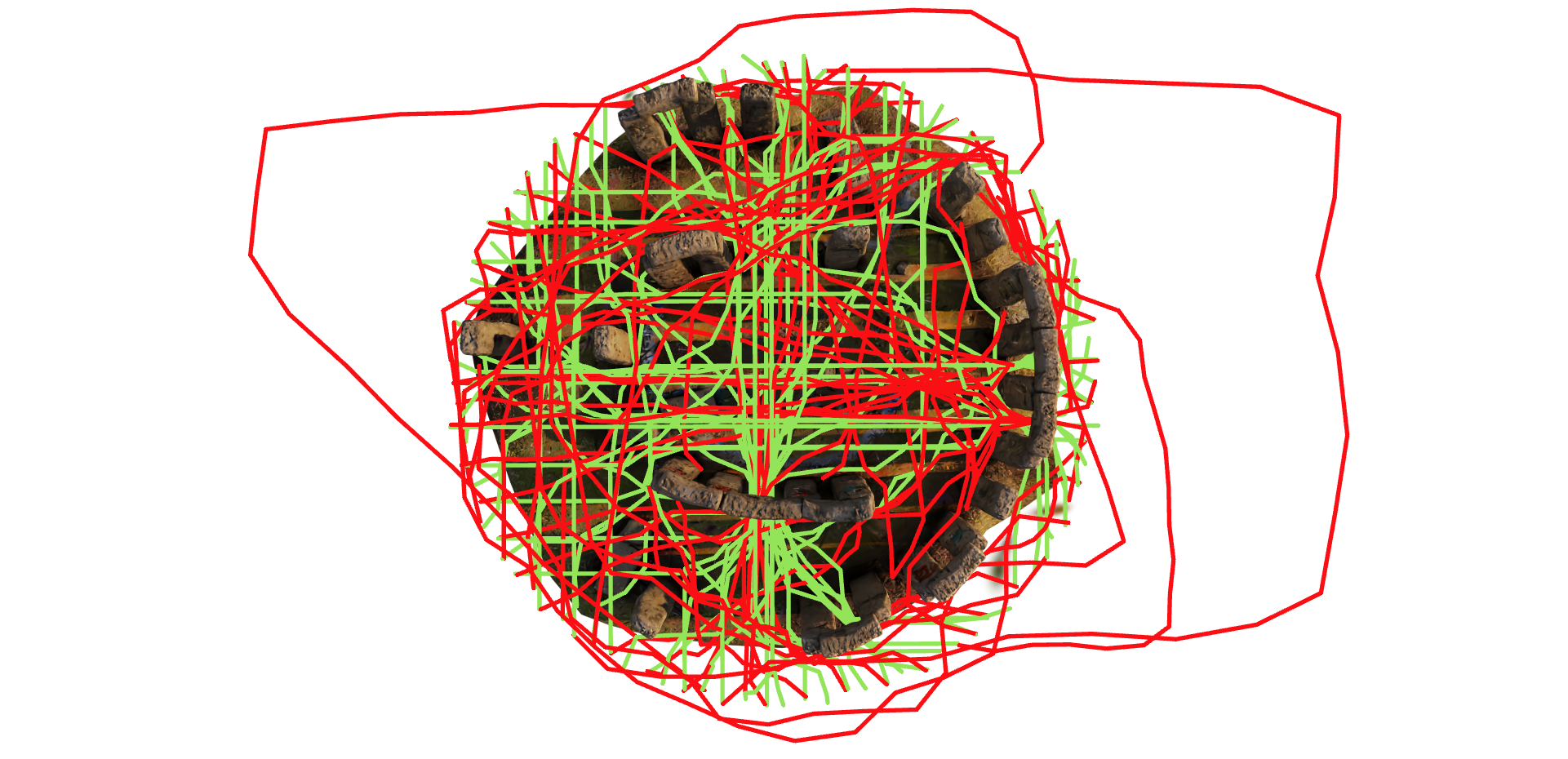}
    \caption{The collection of paths used to evaluate different methods with various start and end goals around Stonehenge. The RRT* path is shown in \textcolor{red}{red} and Splat-Nav~\cite{Chen2024} in \textcolor{green}{green}.}
    \label{fig:stonehenge_paths}
    \vspace{-0.5cm}
\end{figure}

In comparisons with similar methods (\added{Table~\ref{tab:stonehenge}}, Fig.~\ref{fig:stonehenge_paths})\added{, FOCI needs on average less than one second to compute paths that are shorter than those RRT* returns after its fixed planning budget of $100\,\si{\s}$}, while having {similar time performance to state-of-the-art \ac{3dgs} methods such as Splat-Nav}. 
Note that, due to FOCI's ability to model the robot as a collection of Gaussians and therefore consider the robot's orientation, {the smaller safety corridor allowed for shorter paths, leveraging the robot's geometry}.

\begin{table}[t]
	\centering
	\scriptsize
	\setlength{\tabcolsep}{2pt}
	\caption{\added{Planning time and normalized path length of FOCI and the baseline methods for the ANYmal robot. Entries marked with - are not applicable. $N$/$M$: number of environment/robot Gaussians. Mean over the start/goal pairs shown in Fig.~\ref{fig:stonehenge_paths}. $^\dagger$Fixed planning budget, not a time to solution.}}
	\label{tab:stonehenge}
	\begin{tabular}{clcclcc}
		\toprule
		\added{\textbf{Robot}} & \textbf{Environment} & \textbf{N} & \textbf{M} & \textbf{Method} & \textbf{\begin{tabular}[c]{@{}c@{}}Total Comp.\\Time (s)\end{tabular}} & \textbf{\begin{tabular}[c]{@{}c@{}}\added{Path}\\\added{Length (-)}\end{tabular}} \\
		\midrule
		\multirow{3}{*}{\rotatebox{90}{ANYmal}} & \multirow{3}{*}{Stonehenge} & \multirow{3}{*}{138k} & 3 & FOCI & $0.78$ & \textbf{0.88} \\
		&  &  & - & RRT* & \added{$100.0^\dagger$} & $0.92$ \\
		&  &  & - & Splat-Nav & \textbf{0.76} & $1.01$ \\
		\bottomrule
	\end{tabular}
	\vspace{-0.3cm}
\end{table}

\section{Limitations}

\added{
Three current limitations of the algorithm include the following ones:}

 \added{
\textit{a)} Since the jerk and obstacle costs are additive rather than constrained, the optimizer can trade off collision avoidance for lower jerk, occasionally yielding colliding trajectories,
as also observed in NeRF-based planning~\cite{adamkiewicz2022vision}. This could be mitigated by constraining trajectories near the initial guess or retuning the cost weights.}

 \added{
\textit{b)} Velocity, acceleration, and jerk are scaled by a time-regulation factor $\frac{ds}{dt}$ derived from the length of the initial guess. Thus, longer trajectories of the same shape incur larger jerks, letting the jerk cost dominate over obstacle avoidance. Introducing execution time as an explicit decision variable would address this.}

 \added{
\textit{c)} The overlap integral implicitly assumes that high Gaussian density corresponds to solid geometry, which is not always true: \ac{3dgs} also increases density in high-detail regions (edges, fine strands, text), slightly inflating the collision cost there. Moreover, interior density is not guaranteed, so a trajectory that penetrates a surface can get stuck in a local minimum with no gradient to escape. Both issues could be mitigated through more geometry-aware \ac{3dgs}~\cite{chao2024texturedgaussiansenhanced3d, goel2025distance}.}

\added{Finally, the CHOMP and STOMP results should be interpreted with care: both are our own implementations adapted to the Gaussian overlap cost. We expect them to improve with a distance-field hinge cost with an explicit clearance margin, as in their original design, with per-time-step weighting and a finer discretization for STOMP, and with batched GPU evaluation to reduce solve times.}

\vspace{-0.1cm}
\section{Conclusion}

In this work, we proposed a novel \added{robot-agnostic} collision formulation for \acl{3dgs}, that is computed in an efficient parallel manner, and integrated it into a trajectory optimization pipeline. 
We show that it can be effectively used for orientation-aware planning and verify it on the ANYmal quadruped robot, \added{as well as in simulation on a UR5 manipulator and two UAVs.} 
Because we exclusively operate in 3D Gaussian space, representing the environment and the robot as Gaussians, our method can be freely combined with new developments from the 3DGS community. 
Furthermore, we show that this method works on realistic data including scenes captured using the onboard sensor of the robot itself.

\section*{\added{Disclosure of AI use}}
\added{Claude (Anthropic) was used as a coding assistant for part of the software implementation and for language editing of the manuscript. The authors reviewed all AI-assisted code and text and take full responsibility for the content.}

\bibliographystyle{IEEEtranBST/IEEEtran}
\bibliography{references}

\onecolumn
\newpage

\end{document}